\documentclass[letterpaper, 10 pt, conference]{ieeeconf}  

\IEEEoverridecommandlockouts                              

\usepackage{cite}
\usepackage{amsmath,amssymb,amsfonts}
\usepackage{algorithmic}
\usepackage{graphicx}
\usepackage{textcomp}
\usepackage{xcolor}
\usepackage{hyperref}
\usepackage[ruled,vlined]{algorithm2e}
\usepackage{makecell}
\usepackage{lipsum}

\title{\LARGE \bf
Dissimilarity-Based Persistent Coverage Control of Multi-Robot Systems for Improving Solar Irradiance Prediction Accuracy \\in Solar Thermal Power Plants
}

\author{%
Haruki Kawase$^{1}$,
Taiga Sugawara$^{1}$,
and A.~Daniel Carnerero$^{1}$%
\thanks{$^{1}$Graduate School of Engineering Science, The University of Osaka, Osaka, Japan.
{\tt\small kawase.haruki.5or@ecs.osaka-u.ac.jp, sugawara.taiga.xxe@ecs.osaka-u.ac.jp, carnerero.daniel.es@osaka-u.ac.jp}}%
}

\begin{document}

\maketitle
\thispagestyle{empty}
\pagestyle{empty}

\begin{abstract}
Accurate forecasting of future solar irradiance is essential for the effective control of solar thermal power plants. Although various kriging-based methods have been proposed to address the prediction problem, these methods typically do not provide an appropriate sampling strategy to dynamically position mobile sensors for optimizing prediction accuracy in real time, which is critical for achieving accurate forecasts with a minimal number of sensors. This paper introduces a dissimilarity map derived from a kriging model and proposes a persistent coverage control algorithm that effectively guides agents toward regions where additional observations are required to improve prediction performance. By means of experiments using mobile robots, the proposed approach was shown to obtain more accurate predictions than the considered baselines under various emulated irradiance fields. 
\end{abstract}


\section{Introduction}
Solar energy has attracted considerable attention as a major renewable energy source. Representative technologies include photovoltaic (PV) power generation and concentrated solar thermal (CST) power generation. The levelized cost of electricity (LCOE) of PV plants is significantly lower than that of CST plants \cite{STKrigingThermosolarForecast}. Nevertheless, CST plants offer a critical advantage in that thermal energy storage (TES) enables electricity generation during nighttime hours \cite{STKrigingThermosolarForecast}. Among CST technologies, solar tower and parabolic trough systems are the most prevalent. In solar tower plants, hundreds to thousands of heliostats reflect sunlight toward a central receiver mounted on a tower, where a heat transfer fluid (HTF) is heated. In parabolic trough plants, parabolic mirrors concentrate sunlight onto a receiver tube, heating an HTF such as oil that flows through the tube.

In parabolic trough plants, the HTF flow rate must be controlled to maximize power generation while keeping the HTF temperature within operational limits. Model Predictive Control (MPC) is a propitious control strategy for this task\cite{MPCParabolicTrough}, as it can explicitly handle constraints while computing optimal control inputs. Such MPC-based control requires accurate predictions of the spatial distribution of direct normal irradiance (DNI). If the HTF flow is not appropriately controlled, parts of the plant may experience overheating or insufficient heating, resulting in inefficient use of solar energy \cite{MobileRobotIrradianceEstimation}. Consequently, accurate short-term DNI forecasting is essential for the optimal operation of parabolic trough CST plants. DNI losses are often quantified using the cloud factor (CF), a scalar variable taking values between 0 and 1, where 1 corresponds to complete DNI loss and 0 represents nominal DNI\cite{MobileRobotIrradianceEstimation}. 

Kriging has become a de facto standard technique for spatial prediction based on limited observations \cite{STKrigingThermosolarForecast, KKFSensorPlacement}. Kriging interpolates spatial data by estimating covariance structures through variogram models and computing weighted averages of observed values \cite{SpatioTemporalKrigingProbeTaxi, TheOriginsOfKriging}. Numerous studies have extended kriging to spatio-temporal settings \cite{NonseparableCovarianceClass, NonseparableCovariance}, and anisotropic spatio-temporal variograms that depend on wind effects have been shown to improve solar irradiance prediction accuracy \cite{STKrigingThermosolarForecast}. However, prediction models that incorporate anisotropy tend to be complex and require ad-hoc treatments. To address this, Kriging using dissimilarity functions has been studied in \cite{Carnerero2023PredictionRegion}, as it allows for easier model construction and expansion. Kernel-Based Kriging (KB-Kriging) \cite{KBKrigingSolarThermal} is a representative example of this class of approaches for irradiance prediction, and improves prediction accuracy by sufficiently utilizing historical observation data. Nevertheless, the authors consider fixed sensor placement, which limits prediction performance. This is because fixed sensors can only observe spatially limited solar irradiance, making it difficult to detect a spatially non-uniform DNI distribution across the plant. Therefore, appropriate sampling strategies need to be developed so that accurate prediction can be carried out with a minimal number of sensors. 

This motivates the use of mobile robotic sensor networks (RSNs) \cite{MobileRobotIrradianceEstimation}. Field estimation with RSNs is closely related to adaptive sampling, and Gaussian processes (GPs)
are frequently employed\cite{AdaptiveSamplingROM, DistributedCoverageTimeVarying}. In particular, the method in \cite{DistributedCoverageTimeVarying} incorporates GP-based prediction uncertainty into a coverage control algorithm to enhance the estimation of time-varying processes. While coverage control is a standard approach for deploying multi-agent RSNs \cite{coveragecontrol}, it aims at convergence to static optimal positions, thereby failing to fully exploit agent mobility for continuous monitoring tasks. To address this limitation, persistent coverage control has been studied in\cite{CoverageInfoDecay, ChargingStationAssignment, PersistentAwarenessCoverage}, where agents continuously move to monitor the environment. This is particularly important for solar irradiance forecasting, since cloud formations are highly dynamic and spatially complex.

This study aims to develop control laws for mobile sensors that are applicable to a KB-Kriging predictor to improve irradiance prediction. We consider a persistent monitoring problem in a parabolic trough CST plant in which multiple mobile agents continuously monitor a two-dimensional spatial domain. For the convenience of model construction in KB-Kriging, we propose a control framework in which the dissimilarity of KB-Kriging is fed back into the control laws of all agents. To achieve this, we first define a KB-Kriging model for solar irradiance prediction using a dissimilarity function. Next, we introduce a dissimilarity map that
represents how dissimilar a prediction point is from the observed data, and incorporate this map to define
a time-varying importance function. Based on the map, we then design a persistent coverage controller for mobile agents. The effectiveness of the proposed method is demonstrated through experiments using mobile robots under virtual irradiance fields. First, we implement solar irradiance prediction under cloudy conditions, comparing three methods: fixed sensors, a baseline persistent coverage control without the dissimilarity map, and the proposed method. Second, additional experiments are conducted by changing the number of agents and climate conditions to evaluate the performance of the proposed algorithm.

The remainder of this paper is organized as follows. Section \ref{sec:problem} describes the plant model and problem formulation. Section \ref{sec:futureprediction} introduces dissimilarity-based solar irradiance prediction. Section \ref{sec:DBPCC} presents the proposed persistent coverage control strategy. Section \ref{sec:numericalExamples} provides experimental results, and Section \ref{sec:conclusions} concludes the paper.

\section{Problem Formulation}\label{sec:problem}

\begin{figure}[t]
    \centering
    \includegraphics[width=1\linewidth]{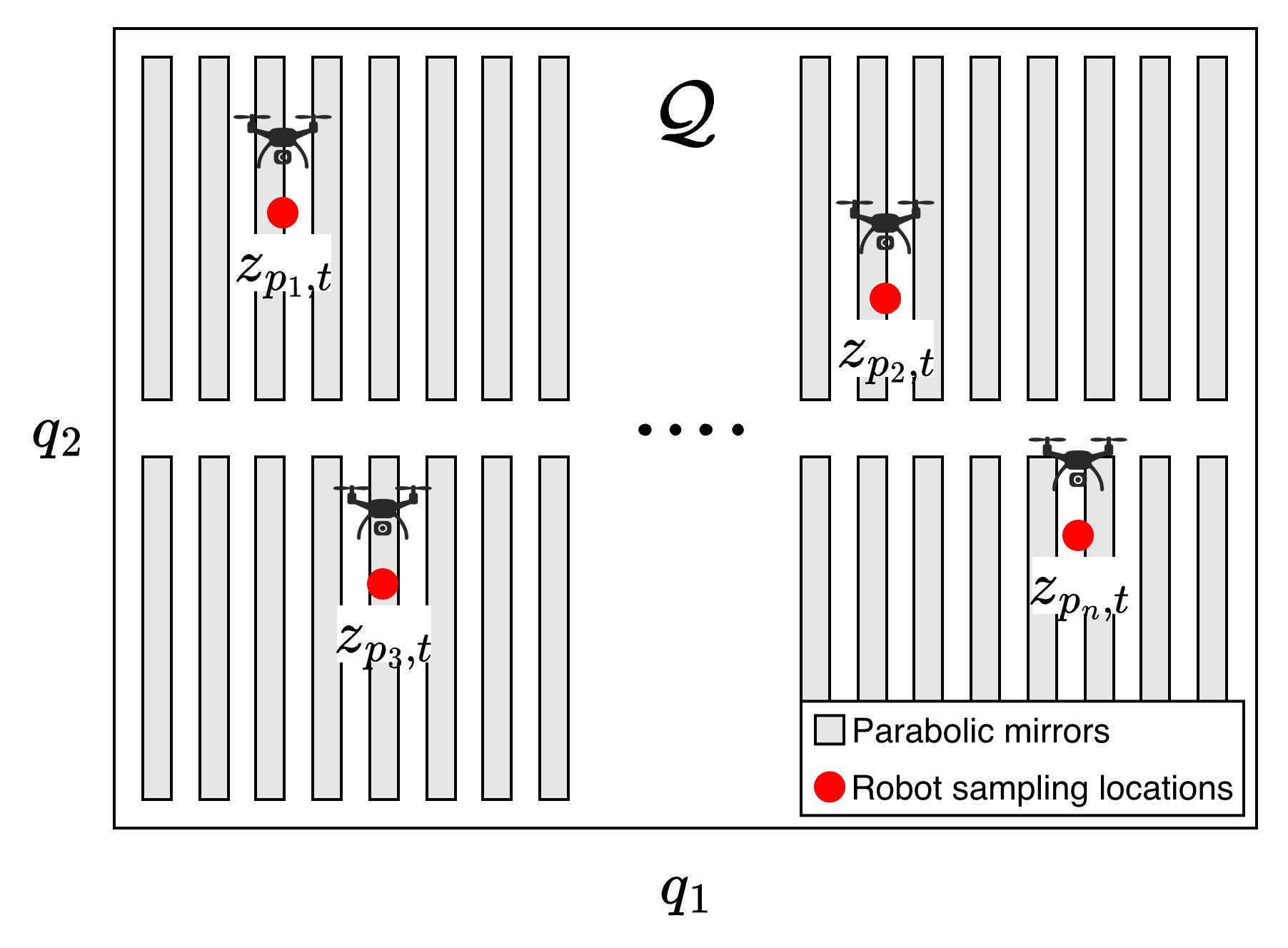}
    \caption{Layout of the plant and spatio-temporal sampling locations of $n$ robots}
    \label{fig:cstplant}
\end{figure}

Let $q = [q_1, q_2]^\top \in \mathcal Q$ denote a vector representing a position in the plant.
Here, $\mathcal Q \subset \mathbb R^2$ is a closed region representing the plant area and is the mission space over which solar irradiance is to be predicted. Let $\mathcal T \subset \mathbb R$ be the time interval during which solar irradiance is observed and predicted, and denote time in this interval by $t \in \mathcal T$.
Fig.~\ref{fig:cstplant} illustrates the plant layout and a team of $n$ robots sampling solar irradiance, where the robots are unmanned aerial vehicles (UAVs) as representative sensing agents. The set of agents $\mathcal A_i$ is defined as

\begin{equation}
    \mathcal A = \{ \mathcal A _ { i } \mid i \in \mathcal I = \{ 1,2,\ldots , n \} \}
\end{equation}
and let $p_i(t) \in \mathcal Q$ denote the position of agent $\mathcal A_i$ at time $t$.
The spatio-temporal vector is defined as $z_{q,t} = [q^\top, t]^\top = [q_1, q_2, t]^\top$.
The true solar irradiance at location $q$ and time $t$ is denoted by $\mathrm{CF}_{q,t}$, and its predicted value by $\hat{\mathrm{CF}}_{q,t}$.
It is necessary to search for the agents’ sampling locations that minimize
\begin{equation}
    E = \frac{1}{t_T - t_0} \int_{t_0}^{t_T} \mathrm{RMSE}(t)\, dt \label{eq:Error},
\end{equation}
where
\begin{equation}\label{eq:rmse}
    \mathrm{RMSE}(t)= \sqrt { \frac { 1 } { | \mathcal Q | } \int_{\mathcal Q} \left|\mathrm{CF}_{q,t} - \hat{\mathrm{CF}}_{q,t} \right|^ { 2 }  d q }.
\end{equation}
Here, \(t_{0}\) and \(t_{T}\) denote the start and end times of the prediction evaluation, respectively, and \(|\mathcal{Q}|\) denotes the area of the region \(\mathcal{Q}\). Problems of this type are typically NP-hard \cite{MultiRobotIPPSparseGP}, and direct minimization of $E$ is not straightforward.
Therefore, we consider a persistent monitoring problem in which an importance function $\phi (q, t) : \mathcal Q \times \mathbb{R} \rightarrow [0,\infty]$ for persistent coverage control \cite{CoverageInfoDecay} is designed to minimize $E$.

\section{Future Irradiance Prediction using Dissimilarity Functions}\label{sec:futureprediction}

\subsection{Dissimilarity Function}
A dissimilarity function quantifies how dissimilar a given element is from a prescribed dataset. Let $\bar d_i \in \mathbb{R}^{n_d}$ denote a sampled data point, and define the dataset as follows.
\begin{equation}
    D = \begin{bmatrix} \bar{d}_1, \ldots, \bar{d}_N \end{bmatrix}
    \in \mathbb{R}^{n_d \times N}
\end{equation}
We consider how dissimilar an arbitrary vector $d \in \mathbb{R}^{n_d}$ is with respect to this dataset.
Here, $n_d$ denotes the data dimension and $N$ the number of sampled data points. The dissimilarity function is defined as
$J :\mathbb{R} ^ { n_d } \times \mathbb{R} ^ { n_d \times N} \rightarrow [ 0, \infty ]$.
A larger value of the dissimilarity function $J(d,D)$ indicates that $d$ is less similar to the dataset $D$, whereas a smaller value indicates higher similarity. A dissimilarity function that is invariant under affine transformations and independent of the coordinate system is given by the following formulation\cite{Carnerero2023PredictionRegion}:
\begin{subequations}\label{eq:dissimilarityfunction}
    \begin{equation}
        J_{\gamma}(d, D) = \underset{\lambda_1,\ldots,\lambda_N}{\operatorname{min}} \; (1-\gamma) \sum_{i=1}^{N} w_{i} \lambda_{i}^{2} + \gamma \sum_{i=1}^{N} |\lambda_{i}| \\
    \end{equation}
    \begin{equation}
    \text{s.t.} \;\;
    \begin{bmatrix}
                D \\
        \mathbf{1}^\top
      \end{bmatrix} \lambda = 
      \begin{bmatrix}
        d \\
        1
      \end{bmatrix}
    \end{equation}
\end{subequations}
This is a convex optimization problem and can be solved efficiently with equality constraints.
$\lambda=[\lambda_1,\ldots,\lambda_N]^\top \in \mathbb R^N$ is a weight vector, and $\gamma \in [0,1)$ is a tuning parameter. The constants $w_i (i=1,\dots, N)$ are weighting coefficients that allow different importance to be assigned to each data point in $D$. This weighting is particularly useful when incorporating local information in nonlinear system modeling\cite{Carnerero2021PIP}. 

\subsection{Kernel-Based Kriging (KB-Kriging)}\label{subsec:KB-Kriging}

We explain Kernel-Based Kriging (KB-Kriging), which computes one-step-ahead prediction $\hat{\mathrm{CF}}_{q , t+ 1}$ using a dissimilarity function \cite{KBKrigingSolarThermal}. From the past $L$ time steps up to time $t$, a total of $N$ observations of $\mathrm{CF}$ are selected, and the observation dataset $Y_t \in \mathbb{R}^{1 \times N}$ is defined as
\begin{equation}
\begin{split}
Y_t = [&\bar{\mathrm{CF}}_{p_1,\,t-L+1},\ldots,\bar{\mathrm{CF}}_{p_n,\,t-L+1},\ldots, \\
&\bar{\mathrm{CF}}_{p_1,\,t},\ldots,\bar{\mathrm{CF}}_{p_n,\,t}]
\end{split}
\end{equation}
In addition, the dataset consisting of the corresponding spatio-temporal vectors, arranged in the same order as $Y_t$, is denoted by $Z_t \in \mathbb{R}^{n_z \times N}$, where $Z_t = [\bar{z}_1,\ldots,\bar{z}_N]$. Here, $n_z$ denotes the dimension of the spatio-temporal vector, which is $n_z = 3$. This dataset is obtained by reindexing
\begin{equation}
\begin{split}
Z_t = [&\bar{z}_{p_1,\,t-L+1},\ldots,\bar{z}_{p_n,\,t-L+1},\ldots, \\
&\bar{z}_{p_1,\,t},\ldots,\bar{z}_{p_n,\,t}]
\end{split}
\end{equation}
Please note that the datasets change at every sampling instant. 

Next, a regressor corresponding to the spatial location and the prediction time is defined as $z_{q,t+1}=[q_1, q_2,t+1]^\top \in \mathbb R^{n_z}$. KB-Kriging employs kernel functions to flexibly represent nonlinear spatial distributions, while requiring the system matrices to be computed only once at each time step. In kernel methods, data with a nonlinear structure are mapped into a high-dimensional Hilbert space through nonlinear operators. Let $\varphi(\cdot):\mathbb R^{n_{z}} \rightarrow \mathbb H$ denote a nonlinear operator, and for simplicity denote $\varphi_z=\varphi(z)$ for $z\in \mathbb{R}^{n_z}$. 
The predicted cloud factor can then be expressed as a weighted average of $Y_t$:
\begin{equation}
    \hat{\mathrm{CF}}_{q,t+1}=Y_t \lambda_{q,t} ^ { * },
\end{equation}
where $\lambda_{q,t}^{*} \in \mathbb R^{N}$ is the optimal weight vector. 
Formally, with $\varphi_{Z_t}=[\varphi_{\bar{z}_1},\ldots,\varphi_{\bar{z}_N}]$, and using the framework of the dissimilarity function \eqref{eq:dissimilarityfunction} with $\gamma = 0$ for mathematical convenience, $\lambda_{q,t}^{*}$ is obtained by solving the following optimization problem. 

\begin{subequations}
    \begin{equation}\label{eq:kbkriging}
        \lambda_{q,t}^{*} = \underset{\lambda_{q,t}}{\operatorname{argmin}}\, \lambda_{q,t}^\top H_{1} \lambda_{q,t} + \left\| \varphi_{Z_t}\lambda_{q,t} - \varphi_{z_{q,t+1}} \right\|^{2}
    \end{equation}
    \begin{equation}
        \text{s.t.}~ \mathbf{1}^\top \lambda_{q,t} = 1
    \end{equation}
\end{subequations}
This problem is obtained by relaxing the linear constraint $Z_t \lambda = z_{q,t+1}$ of Local-Data Kriging \cite{SSKrigingForecastND} and incorporating a kernel-based penalty function to capture nonlinear correlations. Here, $H_1$ is a weighting matrix and is defined as $H_1 = \beta I_N$ using the parameter $\beta$. Expanding the penalty function, the objective function can be rewritten as
\begin{equation}
    \frac { 1 } { 2 } \lambda_{q,t} ^\top H_t \lambda_{q,t} + f_{q,t} ^\top \lambda_{q,t} + 1.
\end{equation}
Following \cite{KBKrigingSolarThermal}, for any pair $z, z' \in \mathbb{R}^{n_z}$, the inner product is defined as $\left\langle \varphi_z, \varphi_{z'} \right\rangle = \varphi_z^\top \varphi_{z'}$, and the following Gaussian kernel with parameters $\sigma$ and $\tau$ is employed:
\begin{equation}
  \left\langle \varphi_{z}, \varphi_{z'}\right\rangle
  = \exp \left\{ -\frac{ \| q - q' \|^2 }{ 2 \sigma^2 } \right\}
  \cdot
  \exp \left\{ -\frac{ \| t - t' \|^2 }{ 2 \tau^2 } \right\}
\end{equation}
Here, $H_t = 2 \left( H_1 + \varphi_{Z_t}^\top \varphi_{Z_t} \right)$ and $f_{q,t}^\top = -2\, \varphi_{z_{q,t+1}}^\top \varphi_{Z_t}$. The kernel-related terms are given by
\begin{equation}
    \varphi_{Z_t}^\top  \varphi_{Z_t}= 
    \scalebox{0.8}{$
    \begin{bmatrix}
\left\langle\varphi_{\bar{z}_1}, \varphi_{\bar{z}_1}\right\rangle & \left\langle\varphi_{\bar{z}_1}, \varphi_{\bar{z}_2}\right\rangle & \ldots & \left\langle\varphi_{\bar{z}_1}, \varphi_{\bar{z}_N}\right\rangle \\
\left\langle\varphi_{\bar{z}_2}, \varphi_{\bar{z}_1}\right\rangle & \left\langle\varphi_{\bar{z}_2}, \varphi_{\bar{z}_2}\right\rangle & \ldots & \left\langle\varphi_{\bar{z}_2}, \varphi_{\bar{z}_N}\right\rangle \\
\vdots & \vdots & & \vdots \\
\left\langle\varphi_{\bar{z}_N}, \varphi_{\bar{z}_1}\right\rangle & \left\langle\varphi_{\bar{z}_N}, \varphi_{\bar{z}_2}\right\rangle & \ldots & \left\langle\varphi_{\bar{z}_N}, \varphi_{\bar{z}_N}\right\rangle
\end{bmatrix}
    $}
\end{equation}

\begin{equation}
\begin{split}
    \varphi_{z_{q,t+1}}^\top \varphi_{Z_t} &= \\
    &\scalebox{0.8}{$
    \begin{bmatrix}
\left\langle\varphi_{z_{q,t+1}}, \varphi_{\bar{z}_1}\right\rangle & \left\langle\varphi_{z_{q,t+1}}, \varphi_{\bar{z}_2}\right\rangle & \ldots & \left\langle\varphi_{z_{q,t+1}}, \varphi_{\bar{z}_N}\right\rangle
\end{bmatrix}
    $}
\end{split}
\end{equation}
This kernel function assumes separability between spatial and temporal components, which implies that there are no spatial and temporal correlations. Also, this kernel function shows the best prediction accuracy for this particular problem in \cite{KBKrigingSolarThermal}. Although the spatio-temporal structure of the kernel can be reconsidered to further enhance accuracy, this study focuses on improving prediction performance through the sampling strategy.

As a result, the dissimilarity function incorporating kernel functions,
$J_\varphi:\mathbb{R} ^ { n_z } \times \mathbb{R} ^ { n_z \times N } \rightarrow [ 0, \infty ]$, is formulated as follows:
\begin{subequations}\label{eq:kbk_dissimilarity}
    \begin{equation}
        J_\varphi\left( z_{q, {t + 1 }} , Z_t \right) = \min_{\lambda_{q,t}} \; \frac { 1 } { 2 } \lambda_{q,t} ^\top H_t \lambda_{q,t} + f_{q,t} ^\top \lambda_{q,t} + 1
    \end{equation}
    \begin{equation}
        \text{s.t.} \; \mathbf{1}^\top \lambda_{q,t} = 1
    \end{equation}
\end{subequations}
The optimal weight vector $\lambda_{q,t}^*$ is the minimizer of
\eqref{eq:kbk_dissimilarity}.
The predicted value is then given by
\begin{equation}\label{eq:kbk_estimation}
\hat{\mathrm{CF}}_{q,t+1} = Y_t \lambda_{q,t}^{*}.
\end{equation}
This dissimilarity function quantitatively measures the dissimilarity between
the prediction regressor $z_{q,t+1}$ and the observed dataset $Z_t$ in the feature space.

\section{Dissimilarity-Based Persistent Coverage Control}\label{sec:DBPCC}
In this section, based on the theory of persistent coverage control with information decay \cite{CoverageInfoDecay}, we propose a method for designing an importance function and a corresponding controller that are suitable for kriging-based prediction.

\subsection{Agent and Information Model}
\paragraph{Agent Model}
Let $u _ { i }(t) \in \mathbb R^2$ denote the control input applied to agent $\mathcal A_i$ at time $t$. Each agent is assumed to follow the following kinematic model:
\begin{equation}
    \dot{p} _ { i }(t) = u_i(t)
\end{equation}
The communication network among agents is assumed to be fully connected, such that information can be shared between any pair of agents $\mathcal{A}_i$ and $\mathcal{A}_j \ (i, j \in \mathcal{I})$.
\paragraph{Measurement Function}
The measurement function represents the sensing capability of an agent as a function of its position. Let $r$ and $C$ denote the sensing radius and a positive constant, respectively, and define $s _ { i }  = \| q - p _ { i }(t) \| ^ { 2 }$. The measurement function is defined as
\begin{align}
    M_{i}(s_{i}) = 
    \begin{cases}
        \dfrac{C}{r^{4}}(s_{i} - r^{2})^2, & \text{if } s_{i} \leq r^{2}, \\
        0, & \text{if } s_{i} > r^{2}.
    \end{cases}
\end{align}
The measurement map is defined as the sum of the measurement functions of all agents:
\begin{equation}
    M(s)=\sum_{i \in \mathcal I} M_i(s_i),
\end{equation}
where $s$ denotes the set of all $s_i$.
\paragraph{Information Model}
The information map is defined as $I ( q , t ) : \mathcal Q \times \mathbb R \rightarrow \mathbb{R} ^ { + }$. The information dynamics are described by the following differential equation using a decay coefficient $\delta \leq 0$:
\begin{equation}
    \frac { \partial } { \partial t } I ( q , t ) = \delta I ( q , t ) + M ( s ).
\end{equation}
The first term represents natural information decay, while the second term represents information acquisition based on the measurement map. Information increases in regions close to the agents and decreases elsewhere. Let $I_{\mathrm{ref}}(q)$ denote the desired information level over the field. The agents are expected to move persistently to achieve the desired information level at all locations.

\subsection{Importance Function Based on Dissimilarity Map}\label{subsec:importance}
Let $\phi ( q , t ) : \mathcal Q \times \mathbb{R} \rightarrow [0,\infty]$ denote a time-varying function that represents the importance of location $q$ at time $t$. This function is used to manage regions of interest for monitoring. In this study, dissimilarity information is incorporated into the importance function design. Specifically, an importance function dependent on the dataset $Z_t$ is constructed as
\begin{equation}
    \phi_{Z_t}( q, t ) = J_\varphi\left( z_{q, {t + 1 }} , Z_t \right),
\end{equation}
which is referred to as the dissimilarity map. Consequently, regions with high kriging dissimilarity are assigned higher importance values. The dissimilarity map is constructed online in a centralized manner based on Section~\ref{subsec:KB-Kriging}. 

\subsection{Control Objective}

The control objective is to persistently move the agents such that no region in the mission space $\mathcal Q$ remains uncovered within the sensing domain, with particular emphasis on regions exhibiting high dissimilarity. Let $e_I = I_{\mathrm{ref}}(q) - I(q,t)$ denote the information error, and define the information penalty function as $h(e_I) = \bigl(\max(0, e_I)\bigr)^2$. Then, the objective function to be minimized is given by
\begin{align}
\min \ H(t) = \int_{\mathcal{Q}} h(e_I)\,\phi_{Z_t}(q,t)\, dq.
\end{align}

\subsection{Dissimilarity-Based Persistent Coverage Control}\label{subsec:proposed_contoroller}

Following \cite{CoverageInfoDecay}, we propose the following controller to achieve the control objective:
\begin{equation}\label{equation:controller1}
    u_i (t) =- k \int _ { q \in \mathcal{S} _ { i } } h ^ { \prime } (e_I) M ^ { \prime } ( s _ { i } ) ( q - p_ { i } ) \phi_{Z_t}( q, t ) \, d q,
\end{equation}
where $\mathcal S _ { i } ( p _ { i }(t) ) = \{ q \in \mathcal Q \mid \| q - p _ { i }(t) \| \leq r \}$ denotes the sensing region, $h ^ { \prime } ( e_I ) = \max(0, 2e_I)$, and $k \in \mathbb{R}^+$ is a control gain.

If the information level within the sensing region exceeds the desired level and no penalty is present, the control input given in \eqref{equation:controller1} becomes zero. This condition can be expressed as
$$
I \geq I_{\mathrm{ref}} \ \forall q \in \mathcal S _ { i }
\;\Rightarrow\;
h' = 0 \ \forall q \in \mathcal S _ { i }
\;\Rightarrow\;
u_i(t) = 0.
$$
When the above condition holds, an alternative controller is employed. Let $\hat{k} \in \mathbb{R}^+$ denote a control gain, and let $\hat{q}_i$ be an arbitrary point within the sensing region $\mathcal S_i$. The controller is defined as
\begin{equation}\label{equation:controller2}
\begin{split}
    \hat{u}_i (t) &=- \hat{k}(p_i(t) - \hat{q}_i).
\end{split}
\end{equation}
To simplify, let us denote the control law in this manner:
\begin{equation}\label{eq:proposedController}
u_i^{*}(t)=
\begin{cases}
u_i(t), & \text{if } h' \neq 0 \text{ for some } q \in \mathcal S_i,\\
\hat{u}_i(t), & \text{if } h' = 0 \ \forall q \in \mathcal S_i.
\end{cases}
\end{equation}
For a sufficiently large gain $k$ such that
$\frac{\partial \phi_{Z_t}}{\partial t}(q,t) \ll k$ and
$\frac{\partial^{2} \phi_{Z_t}}{\partial t^{2}}(q,t) \ll k$, an analysis can be performed to establish the boundedness of the objective function.
Under these conditions, it has been shown that the controller $u_i(t)$ keeps the objective function $H(t)$ bounded as
$0 \le H(t) \le \bar{H}$ for a positive constant $\bar{H}$~\cite{CoverageInfoDecay}.

\section{Experimental Results}\label{sec:numericalExamples}
\subsection{Experimental Validation}\label{subsec:test_items}

In this section, experimental results of one-step-ahead solar irradiance prediction over the mission space $\mathcal{Q}$ using a KB-Kriging predictor are presented, comparing three methods: (A) a fixed sensor deployment, (B) a baseline persistent coverage control method with $\phi(q,t)=1$, and (C) the proposed method. Whereas prediction using (A) the fixed method is conducted entirely
in MATLAB, (B) and (C) are implemented in a hardware-in-the-loop configuration using TurtleBot3 to provide a proof of concept. See \cite{Turtle} for the specifications of TurtleBot3. Although the proposed algorithm is formulated for holonomic systems such as UAVs, it remains effective for nonholonomic systems like TurtleBot3 because the computed control input is used as a high-level reference input for a low-level controller. The number of agents is $n=4$, and we set the prediction interval from $t_0=1$ to $t_T=100$. Prediction is performed sequentially for each time step $t \in \{1,2,\ldots,100 \}$ based on data available at $t-1$.
A PC with Ubuntu 20.04 equipped with intel Core i7 CPU, an NVIDIA GeForce GTX 1650 Mobile/Max-Q, and 16 GB of RAM was used, and TurtleBot3 was controlled using ROS Noetic.

Regarding (A), assuming a uniform importance distribution over the mission space $\mathcal{Q}$, the fixed sensor positions are set to the centroids of the Voronoi regions computed using Lloyd's algorithm. The fixed sensor configurations and initial mobile agent positions are summarized in Table~\ref{tab:init}.

\begin{figure}[b]
    \centering
    \includegraphics[width=1\linewidth]{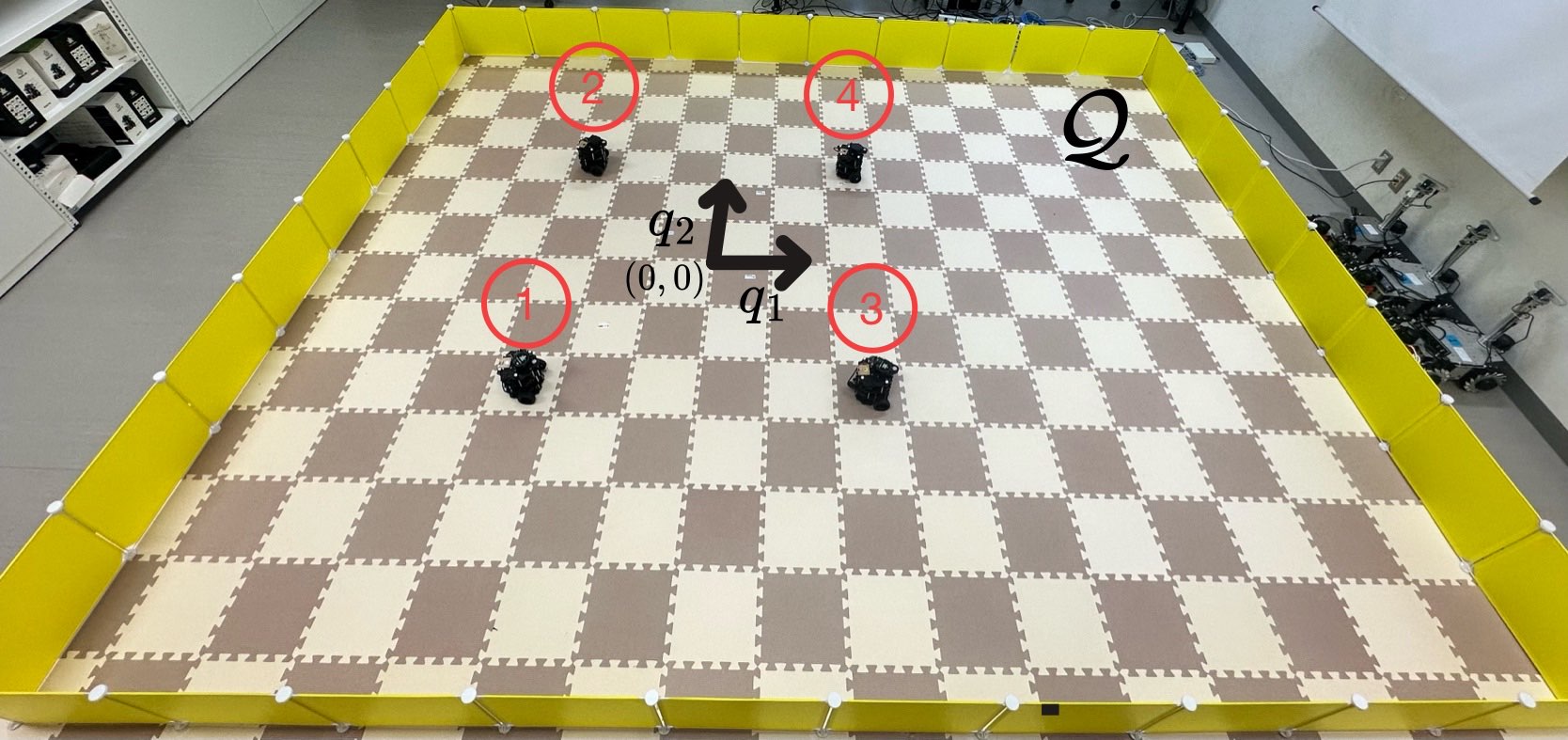}
    \caption{Layout of the experiment room and the initial configurations of 4 mobile agents. }
    \label{fig:room}
\end{figure}

We use the cloudy-condition solar irradiance dataset over $\mathcal{Q}$ provided by J.~G.~Martin et al.~\cite{STKrigingThermosolarForecast}.
The dataset consists of a $97 \times 72$ spatial grid and 2000 time steps. The mission space $\mathcal Q$ is $q_1 \in [-1.41, 2.38], q_2 \in [-1.26, 1.53]$ [m], and is discretized into a $97 \times 72$ grid according to the dataset shape. Continuous spatial integrals are approximated by discrete summation. Note that irradiance measurements can only be obtained at the agent position $p_i(t)$ regardless of $r$. Table~\ref{tab:settings} summarizes the key experimental settings, and the layout of the experiment room can be seen in Fig.~\ref{fig:room}.

\begin{table}[b]
\centering
\caption{Configurations of fixed sensors and mobile agents}
\begin{tabular}{ll}
\hline
\multicolumn{2}{c}{\textit{Fixed sensor configuration}} \\
\hline
\makecell[l]{Sensor positions\\
(Voronoi centroids)}
& $\begin{aligned}
p_1 &= (1.42,\,0.84)\\
p_2 &= (1.45,\,-0.57)\\
p_3 &= (-0.45,\,-0.57)\\
p_4 &= (-0.48,\,0.84)
\end{aligned}$ \\[2mm]
\hline
\multicolumn{2}{c}{\textit{Mobile agent configuration}} \\
\hline
Initial positions 
& $\begin{aligned}
p_1(0) &= (-0.54,\,-0.54)\\
p_2(0) &= (-0.54,\,0.86)\\
p_3(0) &= (0.86,\,-0.52)\\
p_4(0) &= (0.80,\,0.86)
\end{aligned}$ \\[2mm]
\hline
\end{tabular}
\label{tab:init}
\end{table}

\begin{table}[b]
\centering
\caption{Experimental settings}
\begin{tabular}{lc}
\hline
Item & Setting \\
\hline
$\mathcal{Q}$ & \begin{tabular}[t]{c}
$-1.41\leq q_1 \leq 2.38$\\
$-1.26\leq q_2 \leq 1.53$\\
$(97 \times 72 \ \text{grid})$
\end{tabular} \\
$L$ & 10 \\
$t_0,t_T$ & 1,100 \\
$n$ & 4 \\
$C$ & 0.3 \\
$r$ & 0.5 \\
Observation noise 
& Not considered \\
Initial information map 
& $I(q,0)=0\;\forall q$ \\

Reference information map 
& $I_{\mathrm{ref}}(q)=1\;\forall q\in\mathcal Q$ \\
\hline
\end{tabular}
\label{tab:settings}
\end{table}

Self-localization is performed using odometry, and coordinate frame transformations are managed by the ROS tf library.
Pose information is obtained at a frequency of 60 Hz, and the maximum velocity and acceleration of the robots are limited to 0.15 m/s and 0.1 m/s$^2$, respectively.

Algorithm~\ref{algorithm:1} summarizes the computation of 
the next sampling locations and velocity commands of the robots. At each time step, a virtual input $u_i^{*}(t)$ is calculated according to \eqref{eq:proposedController}.
The next target position $p_i^{\mathrm{next}}$ is periodically updated as
$p_i^{\mathrm{next}}=p_i(t)+u_i^{*}(t)$,
and is distributed from the goal node via the ROS topic communication (\texttt{/tb3\_i/goal}). Let $p_i^{r}=p_i^{\mathrm{next}}-p_i$ denote the deviation between the current position and the next position. In addition, let $\theta_i$ and $\theta_i^{\mathrm{next}}$ be the current and the desired heading angle of robot $i$ to the next location, respectively. The angular deviation is defined as $\theta_i^{r} = \theta_i^{\mathrm{next}} - \theta_i$. To guide the robot toward the next target position, the linear and angular velocity inputs $v_i,\omega_i$ are determined by the following control laws:
\begin{equation}
\begin{aligned}
v_i &= K_p\,p_i^{r},\\
\omega_i &=
\begin{cases}
 K_\omega\,\theta_i^{r}, & v_i > 0,\\
- K_\omega\,\theta_i^{r}, & v_i \le 0,
\end{cases}
\end{aligned}
\label{eq:vel}
\end{equation}
where $K_p=0.8$ and $K_\omega = 1.0$. The update of the next sampling location $p_i^{\mathrm{next}}$ is executed every 8~s, whereas the velocity commands $v_i(\cdot)$ and $\omega_i(\cdot)$ are transmitted at 10~Hz via a wireless LAN router using the most recently received target position. In case the robots come close to each other, collision avoidance is implemented by incorporating a repulsive function into the velocity commands. Note that the solar irradiance observations are obtained from a virtual environment for (B) and (C). Although the typical forecasting cycle for solar irradiance in solar thermal power plants ranges from 60 s to 300 s \cite{STKrigingThermosolarForecast}, the update interval of 8~s is adopted here for experimental validation and does not aim to reflect an actual operational forecasting cycle.

The tuning parameters are $\beta, \sigma, \tau$ for (A), and $k, \hat{k}, \beta, \sigma, \tau, \delta$ for (B) and (C).
These parameters are optimized using MATLAB’s \texttt{fmincon} to minimize $E$ in~\eqref{eq:Error}, using 100 data points from the first half of the dataset.
The tuning results are summarized in Table~\ref{tab:tuning}.
For prediction accuracy evaluation, another 100 data points extracted from the second half of the dataset are used. 

\begin{algorithm}[t]
\caption{Computation of sampling locations and control inputs}
\label{algorithm:1}
\textbf{Initialize} $p_i(0)$ for all $i \in \mathcal I$\;
\For{$t = 0,1,2,\dots,t_T-1$}{%
\tcp{Path planning and prediction step (every 8 s)}
  \For{\textnormal{each robot} $i \in \mathcal I$}{%
    Compute reference input $u_i^{*}(t)$ in \eqref{eq:proposedController}\;
    $p_i^{\mathrm{next}} \gets p_i(t) + u_i^{*}(t)$\;
    Publish $p_i^{\mathrm{next}}$ via ROS topic \texttt{/tb3\_i/goal}\;
    \While{\textnormal{receiving} $p_i^{\mathrm{next}}$}{%
    \tcp{Control loop (10 Hz)}
      Obtain current pose $p_i$ and $\theta_i$ by odometry\;
      $p_i^r \gets p_i^{\mathrm{next}} - p_i$\;
      $\theta_i^r \gets \theta_i^{\mathrm{next}} - \theta_i$\;
      Compute $v_i$ and $\omega_i$ in \eqref{eq:vel} and send them to robot\;
    }
  }
}
\end{algorithm}

\begin{table}
\centering
\caption{Tuning parameters}
\label{tab:tuning}
\begin{tabular}{lccc}
\hline
Parameter & (A) Fixed & (B) Baseline & (C) Proposed \\ \hline
$\beta$        & 0.0003665 & 0.211844 & 0.169103 \\
$\sigma$       & 0.297397  & 0.166996 & 0.202815 \\
$\tau$         & 0.119574  & 0.303474 & 0.329897 \\
$k$            & --        & 0.016427 & 0.057800 \\
$\hat{k}$      & --        & 0.268257 & 0.399603 \\
$\delta$       & --        & -0.138640 & -0.209257 \\ \hline
\end{tabular}
\end{table}

\begin{figure}[t]
    \centering
    \includegraphics[width=1\linewidth]{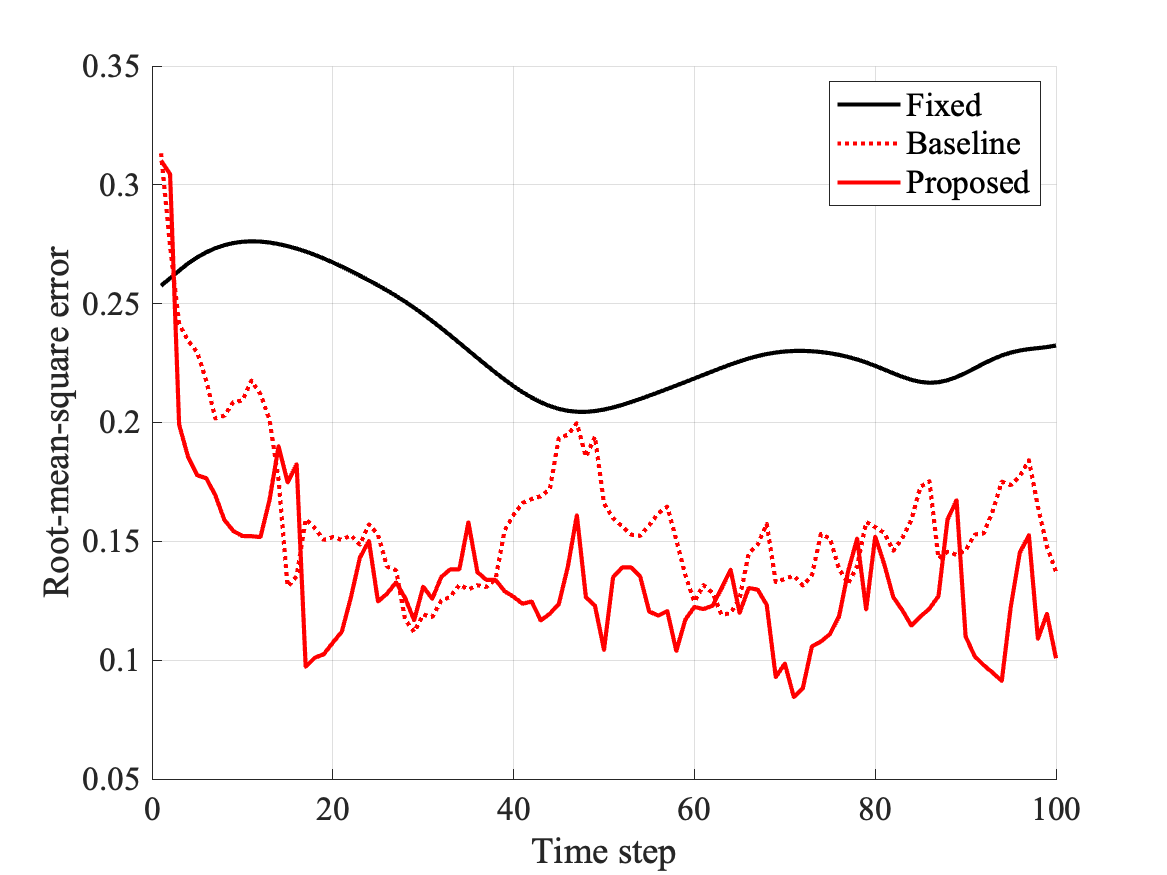}
    \caption{Time-series RMSE comparison (n=4) among (A) the fixed method, (B) the baseline method, and (C) the proposed method.}
    \label{fig:mean_rmse}
\end{figure}

Fig.~\ref{fig:mean_rmse} shows the time-series RMSE for each method.
(C) maintains lower RMSE than the other methods at almost all time steps, confirming improved accuracy.
As shown in Table~\ref{tab:error_comparison}, the time-averaged RMSE $E$ for the fixed, baseline, and proposed methods indicates that (C) achieves the best performance.
Experiment videos are available \href{https://www.youtube.com/playlist?list=PLQ2skd3uwUmF7x0B8L1orHAf1PHioka3S}{here}.

\begin{table}[t]
\centering
\caption{Error comparison ($n=4$) under cloudy weather}
\begin{tabular}{lccc}
\hline
 & (A) Fixed & (B) Baseline & (C) Proposed \\
\hline
$E$ & 0.235 & 0.160 & \(\mathbf{0.134}\) \\
\hline
\end{tabular}
\label{tab:error_comparison}
\end{table}

In (A), the coordinate information of $Z_t$ stored in the dataset is fixed over time, and accordingly, the observations in $Y_t$ are limited to solar irradiance data at those fixed coordinates.
When predicting the solar irradiance at an arbitrary coordinate $q$ using the regressor $z_{q,t+1}=[q_1, q_2, t+1]^\top$, the predicted value is given by $\hat{\mathrm{CF}}_{q,t+1} = Y_t \lambda^*$ according to~\eqref{eq:kbk_estimation}.
However, if the prediction point $q$ is far from the observed coordinates in $Z_t$ in feature space, there are many prediction points that lie outside the convex hull defined by the observations. This leads to a large dissimilarity value and degraded prediction performance. Conversely, if $q$ lies near one of the coordinates in $Z_t$ in the feature space, neighboring information can be exploited, resulting in predictions with smaller dissimilarity.
Therefore, in (A), predictions with large dissimilarity dominate in regions without sensor placement, leading to inferior prediction accuracy compared to the other methods.

\begin{figure*}[h]
    \centering
    \includegraphics[width=1\linewidth]{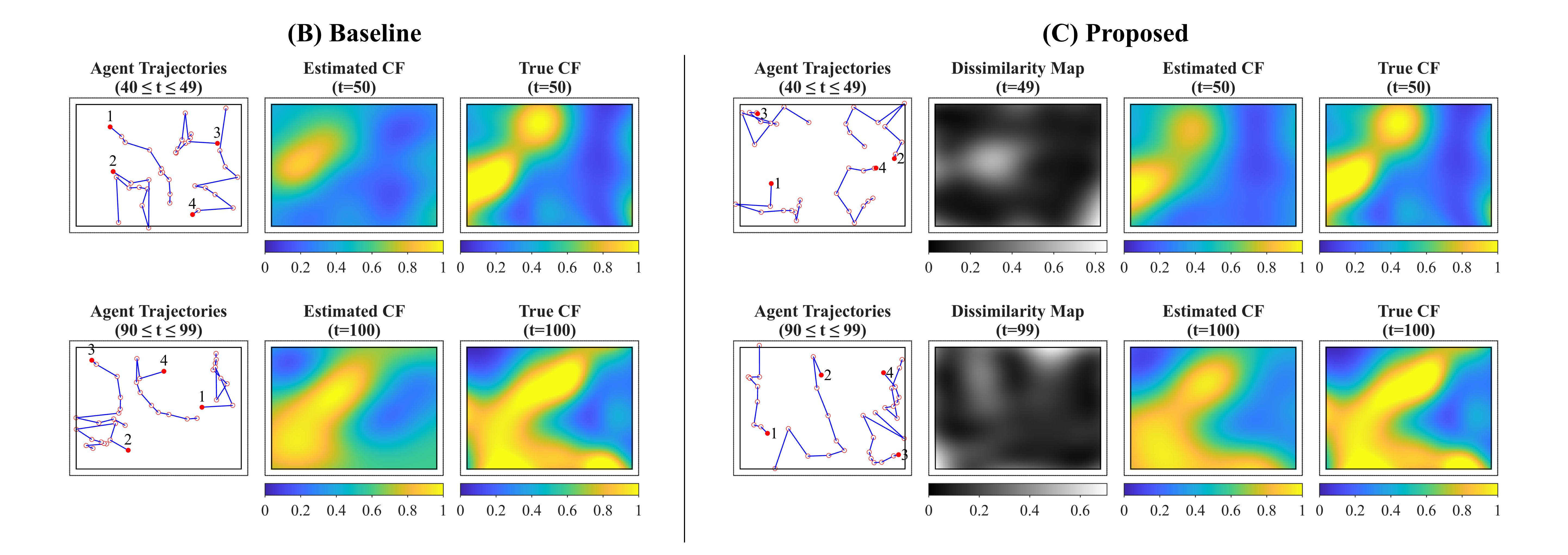}
    \caption{Comparison of the estimation performance between (B) the baseline method (left) and (C) the proposed method (right) corresponding to different time steps (top: $t=50$ and bottom: $t=100$). The inner rectangle represents the region $\mathcal Q$, while the outer rectangle indicates the area extended by 30 cm both horizontally and vertically. The horizontal axis corresponds to $q_1$, and the vertical axis corresponds to $q_2$. Regarding the agent trajectories (blue lines), open red circles denote past sampling points, and filled red circles indicate the current agent positions. A total of $n \times L = 4 \times 10 = 40$ points are sampled in each field. Agents sometimes appear outside the region $\mathcal Q$, which is due to localization errors. (C) incorporates the dissimilarity map into sampling strategies, leading to lower error compared with (B). }
    \label{fig:compare}
\end{figure*}

Fig.~\ref{fig:compare} compares the estimation performance between (B)  and (C), including the agent trajectories for both methods over two time intervals.
In (B), the agents continuously change sampling locations, causing spatial diversity in the coordinates of $Z_t$ and the corresponding observations $Y_t$.
This increases the likelihood of having nearby observations for kriging at arbitrary locations, reducing highly dissimilar predictions compared to (A).
However, because (B) aims to monitor the entire $\mathcal Q$, at certain times, the agents may cluster in similar regions or explore areas where they have already sampled data.
This implies that kriging in unvisited regions relies strongly on highly dissimilar estimates.
In contrast, (C) incorporates regions of high dissimilarity into the sampling decision, driving the agents to move in a way that suppresses the overall dissimilarity map.
As a result, the convex hull characterized by the observed data is expanded, preventing highly dissimilar predictions from dominating solar irradiance estimation across the field. 

\begin{figure}[t]
    \centering
    \includegraphics[width=1\linewidth]{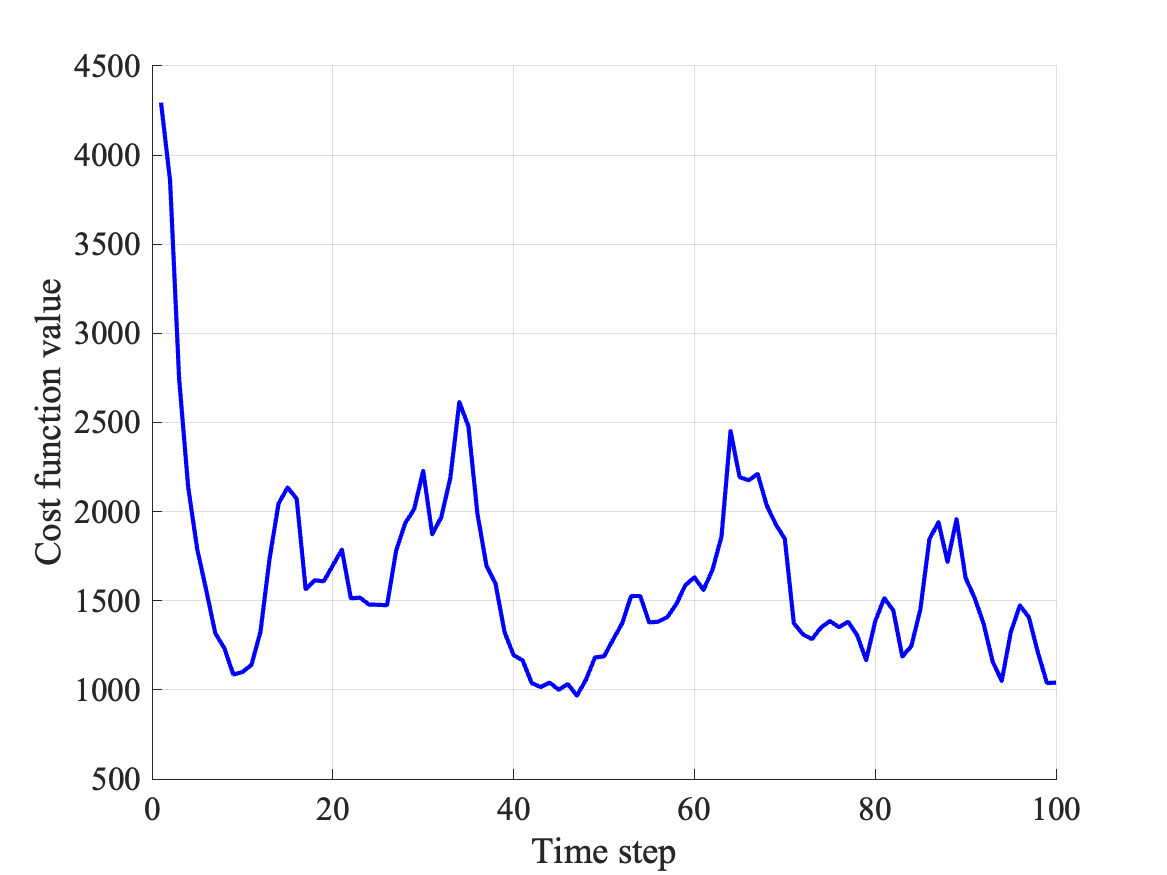}
    \caption{Time-series evolution of the objective function value in (C) the proposed method. }
    \label{fig:cost}
\end{figure}

Fig.~\ref{fig:cost} presents the time evolution of the objective function $H(t)$ for (C).
The values remain bounded, indicating the stability of the control law.
This shows that when dissimilarity increases in regions far from observation points, the agents move toward those regions to reduce dissimilarity. For confirmation, the first- and second-order time derivatives of the dissimilarity map in (C) were numerically approximated. As a result,
$\max_{t}\max_{q}\left| \frac{\partial \phi_{Z_t}}{\partial t}(q,t) \right|
\approx 1.93 \times 10^{-4} \ll k$,
$\max_{t}\max_{q}\left| \frac{\partial^{2} \phi_{Z_t}}{\partial t^{2}}(q,t) \right|
\approx 2.88 \times 10^{-5} \ll k$,
confirming that the necessary assumptions for demonstrating the boundedness of the objective function are satisfied.

\subsection{Performance Evaluation}

To further evaluate the performance of (C), solar irradiance prediction under cloudy conditions is conducted with $n=3,5,6$ agents.
We reuse the tuning parameters shown in Table~\ref{tab:tuning}.
Initial agent positions are randomly selected within the rectangular region
$q_1 \in [-0.54, 0.80], \quad q_2 \in [-0.54, 0.86]$.
All other conditions are identical to those in Section~\ref{subsec:test_items}.
The comparison method is (B) the baseline persistent coverage control.

\begin{figure}[t]
    \centering
    \includegraphics[width=1\linewidth]{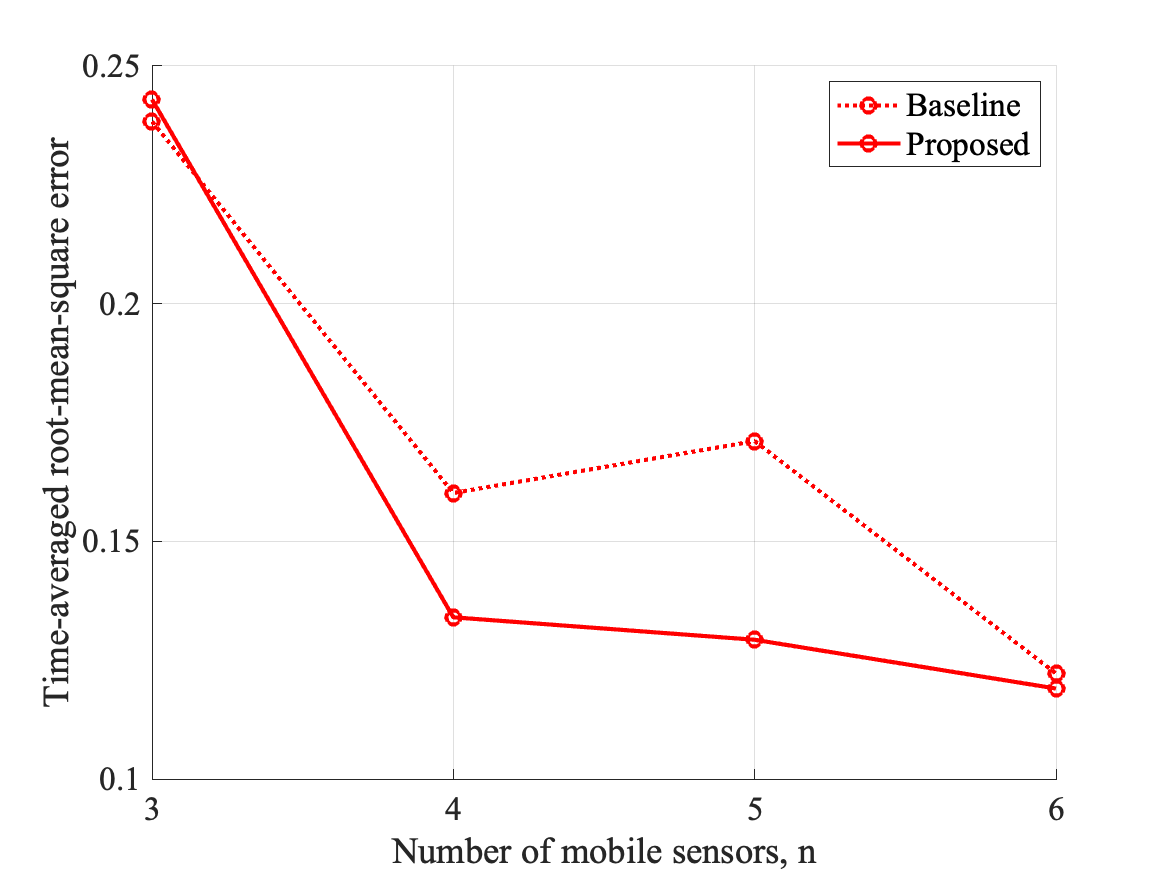}
    \caption{Prediction accuracy with respect to the number of mobile sensors. }
    \label{fig:scalability}
\end{figure}

As shown in Fig.~\ref{fig:scalability}, except for the case $n=3$, (C)  achieves higher prediction accuracy than (B).
For both methods, prediction accuracy tends to improve as the number of sensors increases, since more sampling locations become available.
However, increasing the number of agents introduces trade-offs with system scalability and sensor deployment costs, which are left for future work.
When $n=3$, no significant difference is observed between (B) and (C). This is attributed to the extremely limited number of agents and historical data relative to the large observation area, resulting in uniformly high dissimilarity and making the adaptive sampling ineffective.
When $n=6$, the performance gap narrows because the prediction accuracy saturates with a sufficient number of agents.

To evaluate robustness against climate variation, solar irradiance prediction is performed under three additional weather conditions (standard, sunny, very cloudy) in addition to the cloudy weather.
The parameters in Table~\ref{tab:tuning} are used again, and all other conditions remain the same as in Section~\ref{subsec:test_items}.
As shown in Table~\ref{tab:weather_comparison}, the prediction accuracy of (C) outperforms that of (B) under all weather conditions.
This demonstrates that (C) consistently improves solar irradiance prediction accuracy across varying climates, indicating that the proposed sampling strategy is generic, and even if the weather changes, we do not need to tune the parameters again. 

\begin{table}[b]
  \centering
  \caption{Prediction accuracy under different weather conditions ($n=4$)}
  \label{tab:weather_comparison}
  \begin{tabular}{lcc}
    \hline
    Weather & (B) Baseline $E$ & (C) Proposed $E$ \\
    \hline
    standard    & 0.156 & \textbf{0.142} \\
    sunny       & 0.166 & \textbf{0.120} \\
    cloudy      & 0.160 & \textbf{0.134} \\
    very cloudy & 0.144 & \textbf{0.125} \\
    \hline
  \end{tabular}
\end{table}

\section{Conclusions}\label{sec:conclusions}

In this study, we designed a novel controller that incorporates the dissimilarity map into the importance function of persistent coverage control to improve solar irradiance prediction accuracy in concentrated solar thermal (CST) power plants. The proposed method determines the sampling locations of mobile sensors in an adaptive way, thereby fully exploiting the capability of the KB-Kriging predictor. 
Experimental results using real robots demonstrate that the proposed method significantly reduces prediction errors compared with fixed sensor configurations and the baseline persistent coverage control.

However, the size of commercial CST plants can be up to 780 ha~\cite{SolarFormDetection}, and communication connectivity among agents cannot always be guaranteed. 
Moreover, centralized computation poses scalability issues. 
Therefore, future work needs to focus on developing a distributed framework for both prediction and construction of the dissimilarity map, which are currently performed in a centralized manner.

\section*{Acknowledgment}
The solar irradiance datasets used in this study were provided by J. G. Martin and J. M. Maestre. The authors are indebted to both researchers for generously providing the datasets. 

\end{document}